\documentclass[10pt,twocolumn,letterpaper]{article}

\usepackage[pagenumbers]{cvpr} 

\usepackage{graphicx}
\usepackage{amsmath}
\usepackage{amssymb}
\usepackage{booktabs}
\makeatletter
\@namedef{ver@everyshi.sty}{}
\makeatother
\usepackage{tikz}
\usepackage{float}
\usepackage{array}
\usepackage[accsupp]{axessibility}
\newcolumntype{P}[1]{>{\centering\arraybackslash}p{#1}}
\usepackage[pagebackref,breaklinks,colorlinks]{hyperref}

\usepackage[capitalize]{cleveref}
\crefname{section}{Sec.}{Secs.}
\Crefname{section}{Section}{Sections}
\Crefname{table}{Table}{Tables}
\crefname{table}{Tab.}{Tabs.}

\def\cvprPaperID{9382} 
\def\confName{CVPR}
\def\confYear{2023}

\begin{document}
\newcommand{\sys}{\textsc{DeGPR}}

\title{\sys: Deep Guided Posterior Regularization \\ for Multi-Class Cell Detection and Counting}

\author{
Aayush Kumar Tyagi$^{1*}$, Chirag Mohapatra$^{1*}$, Prasenjit Das$^{3}$,
Govind Makharia$^{3}$,\\  
Lalita Mehra$^{3}$, Prathosh AP$^{2}$, Mausam$^{1}$\\
$^{1}$IIT Delhi ~~~~~~$^{2}$IISc, Bangalore
~~~~~~$^{3}$AIIMS, New Delhi\\
{\small \{tyagiaayushkumar, chirag131020, prasenaiims, govindmakharia,  mehralalita9910, prathoshap\}@gmail.com, mausam@cse.iitd.ac.in
}
}

\maketitle
\def\thefootnote{*}\footnotetext{Equal contribution}

\begin{abstract}
   Multi-class cell detection and counting is an essential task for many pathological diagnoses. Manual counting is tedious and often leads to inter-observer variations among pathologists. 
   While there exist multiple, general-purpose, deep learning-based object detection and counting methods, they may not readily transfer to detecting and counting cells in medical images, due to the limited data, presence of tiny overlapping objects, multiple cell types, severe class-imbalance, minute differences in size/shape of cells, etc. 
   
   In response,  we propose \emph{guided posterior regularization  (\sys)}, which assists an object detector by guiding it to exploit discriminative features among cells. The features may be pathologist-provided or inferred directly from visual data. 
   We validate our model on two publicly available datasets (CoNSeP and MoNuSAC), and on MuCeD, a novel dataset that we contribute. MuCeD consists of 55 biopsy images of the human duodenum for predicting celiac disease. We perform extensive experimentation with three object detection  baselines on three datasets to show that \sys{} is model-agnostic, and consistently improves baselines obtaining up to 9\% (absolute) mAP gains. 
\end{abstract}
\section{Introduction}
\label{sec:intro}
Multi-class multi-cell detection and counting (MC2DC) is the problem of identifying and localizing bounding boxes for different cells, followed by counting of each cell class. MC2DC aids diagnosis of many clinical conditions. For example, CBC blood test counts red blood cells, white blood cells, and platelets, for diagnosing anemia, blood cancer, and infections \cite{george2003understanding,pandey2020target}. 
MC2DC over malignant tumor images helps assess the resistance and sensitivity of cancer treatments \cite{fazio2017fishing}. 
MC2DC over duodenum biopsies is needed to compute the ratio of counts of two cell types for diagnosing celiac disease  \cite{das2019quantitative}. Cell counting is a tedious process and often leads to significant inter-observer and intra-observer variations \cite{corazza2007comparison,ensari2010gluten}. This motivates the need for an AI system that can provide robust and reproducible predictions.


Standard object detection models such as Yolo \cite{glenn_jocher_2022_6222936}, Faster-RCNN \cite{ren2015faster} and EfficientDet \cite{tan2020efficientdet} have achieved state-of-the-art performance on various object detection settings. However, extending these to detecting cells in medical images poses several challenges. These include limited availability of annotated datasets, tiny objects of interest (cells) that may be overlapping, similarity in the appearance of different cell types, and skewed cell class distribution. Due to the non-trivial nature of the problem, MC2DC models may benefit from insights from trained pathologists, e.g., via discriminative attributes. For instance, in duodenum biopsies, intraepithelial lymphocytes (IELs) are structurally smaller, circular, and darker stained, whereas epithelial nuclei (ENs) are bigger, elongated, and lighter. A key challenge lies in incorporating these expert-insights within a detection model. A secondary issue is that such insights may not always be available or may be insufficient -- this motivates additional data-driven features.

We propose a novel deep guided posterior regularization (\sys) framework. Posterior regularization (PR) is an auxiliary loss \cite{ganchev2010posterior}, which enforces that the posterior distribution of a predictor should mimic the data distribution for the given features. 
We call our method deep guided PR, since we apply it to deep neural models, and it is meant to formalize the clinical guidance given by pathologists.
\sys{} incorporates PR over two types of features, which we term explicit and implicit features. Explicit features are introduced through direct guidance by expert pathologists. Implicit features are learned feature embeddings for each class, trained through a supervised contrastive loss \cite{khosla2020supervised}. Subsequently, both features are feed into a Gaussian Mixture Model (GMM). \sys{} constrains the distributions over the predicted features to follow that of the ground truth features, via a KL divergence loss between them.  

We test the benefits of \sys{} over three base object detection models (Yolov5, Faster-RCNN, EfficientDet) on three MC2DC datasets. Of these, two are publicly available:  CoNSeP \cite{graham2019hover} and MoNuSAC \cite{verma2020multi}. We additionally contribute a novel  MuCeD dataset for the detection of celiac disease. MuCeD consists of 55 annotated biopsy images of the human duodenum, which have a total of 8,600 cell annotations of IELs and ENs. We find that \sys{} consistently improves detection and counting performance over all base models on all datasets. For example, on MuCeD, \sys{} obtains a 3-9\%
mAP gain for detection and a 10-35\% reduction in mean absolute error for counting two cell types. 


In summary, 
(a) we propose \sys{} to guide object detection models by exploiting the discriminative visual features between different classes of cells; (b) we use supervised contrastive learning to learn robust embeddings for different cell classes, which are then used as implicit features for \sys{}; (c) we introduce MuCeD, a dataset of human duodenum biopsies, which has 8,600 annotated cells of two types; and (d) we experiment on three datasets, including MuCeD, and find that \sys{} strongly improves detection and counting performance over three baselines.  We release our dataset and code for further research.\footnote{\tt https://github.com/dair-iitd/DeGPR}

\begin{figure}
    \includegraphics[scale = 0.24, width=7.75cm]{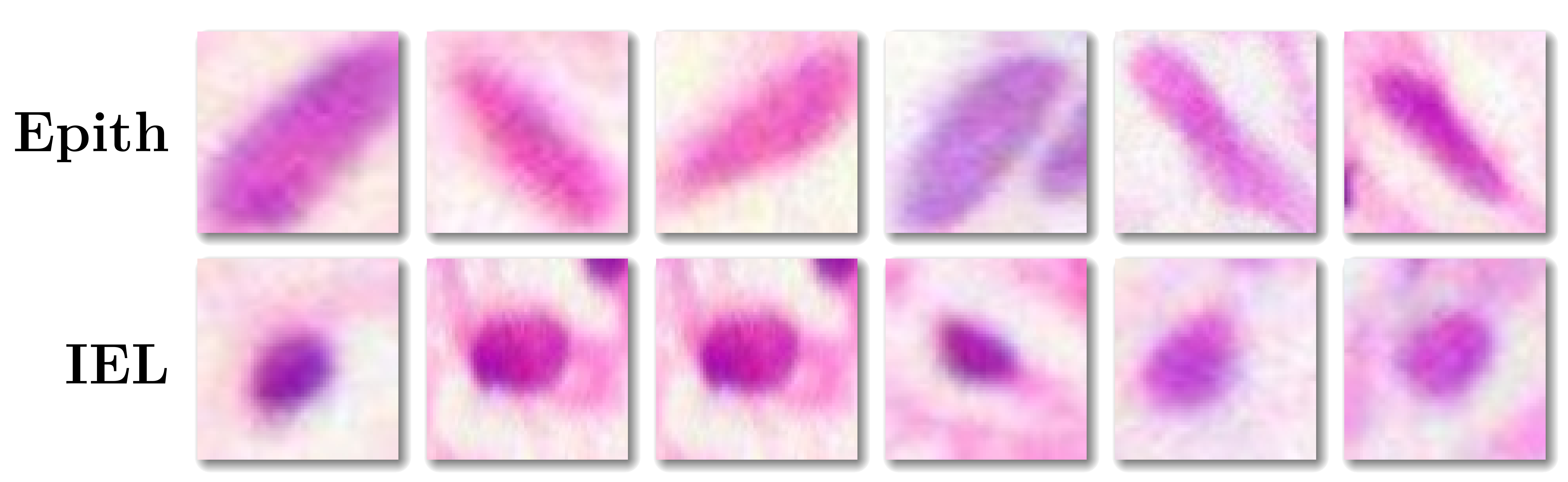}
    \caption{Visual dissimilarities between IELs and ENs. ENs (first row) are lighter stained, bigger and elongated in structure. IELs (second row) are darker stained, smaller, and circular in shape. }
    \label{fig:Shape_size}
\end{figure}

\begin{figure*}[h!]
    \centering
    \includegraphics[scale = 0.23, width = 19.35cm]{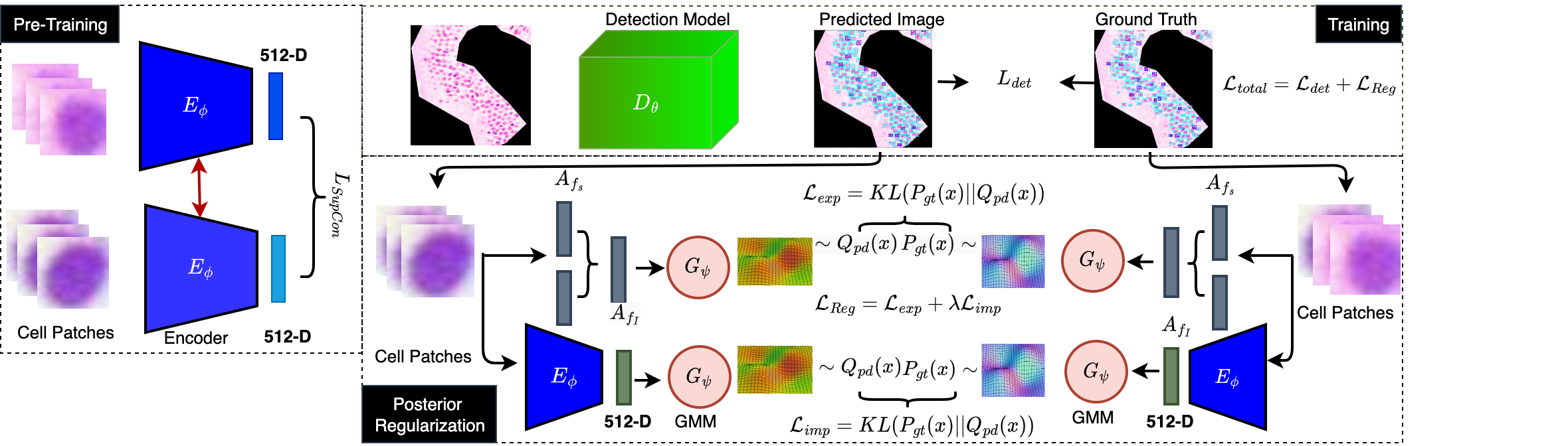}
    \caption{In pre-training stage cell patches are used to pre-train the contrastive encoder to differentiate the cell types. During training step, cell patches are compute based on predicted image from object detection model and ground truth images. These patches are used to compute average size $(A_{f_{s}})$, intensity $(A_{f_{I}})$ and contrastive embedding which will be used as feature vector to train GMM model $(G_{\psi})$. $P_{gt}(x)$ and $Q_{pd}(x)$ are sampled from GMM's and KL divergence is computed which is used to compute $L_{reg}$.}
    \label{fig:Model diagram}
\end{figure*}

\section{Related work}

\noindent
\textbf{Object Detection in Medical Images:}
Object detection is the problem of localization and classification of objects of interest from an image. There are numerous object detection approaches in the literature, such as R-CNN \cite{girshick2014rich}, Yolo \cite{redmon2016you} and RetinaNet \cite{lin2017focal}.  In this work, we experiment with Yolov5 -- the latest in the Yolo series, 
Faster-RCNN -- an improvement over R-CNN, and EfficientDet -- the detection framework built on EfficientNet backbone \cite{tan2019efficientnet}. 


There are two prominent ways of localization over medical images \cite{litjens2017survey}.
In cases where the exact location of an object is not required, detection is done by creating slices of images and subsequently performing classification on each patch \cite{gallego2018glomerulus, cirecsan2013mitosis, tabibu2019pan}. In cases where location is important, standard object detection models are used after fine-tuning on medical datasets \cite{ribli2018detecting, li2019clu, liu2019privacy, jaeger2020retina, li2019weakly}. However, in most medical applications, limited availability of annotated data severely impacts the performance of fine-tuned models \cite{tajbakhsh2016convolutional}.

\vspace{1ex}
\noindent
\textbf{Methods for Cell Detection:}
One common approach for cell detection is to first perform object segmentation, followed by classification.
Segmentation can provide a better solution for the detection task \cite{fujita2020cell}, as it is easier to impose spatial \cite{abousamra2021multi} or geometric \cite{tofighi2019prior, jiang2020geometry} priors over an explicit cell segmentation mask.
At the same time, a pathologist's annotation effort in labeling segmentation masks is significantly higher than annotating bounding boxes. In the spirit of saving annotation effort, our work focuses on cell detection using annotated bounding boxes \cite{cruz2013deep, xu2015stacked}. 
An alternate annotation strategy to bounding boxes, is to annotate centroids \cite{wollmann2021deep, xie2015beyond, sirinukunwattana2016locality} or use attention over feature maps \cite{tomita2019attention, sugimoto2022multi, lei2020attention}. It will be interesting to extend our work to these settings. 



\vspace{1ex}
\noindent
\textbf{Object Detection for Cell Counting:}
 Broadly, there are two main approaches proposed for cell counting in the literature: one is inspired by density-based methods, and the second models counting as a by-product of cell detection. Density-based methods use density maps instead of bounding boxes as labels and evade the hard task of localization  \cite{lempitsky2010learning, paul2017count, sau-net}. 
Existing density-based approaches cannot handle multiple cell types, and hence cannot be directly used for our multi-class cell counting task. In the second approach, counting is generally done as a by product of predicted bounding boxes \cite{dralus2021automatic, albuquerque2021object}. It can also be done over predicted segmentation masks \cite{morelli2021automating}, but the challenge of data annotation for segmentation becomes relevant here too. \sys{} uses counting over predicted bounding boxes, and outperforms natural extensions of density based models.


\section{Methods}

In the problem of multi-class multi-cell detection and counting (MC2DC), we are given an input histopathology image $im$ and the set of $n$ different cell classes $C = \{c_1,c_2,\dots c_n\}$. The goal is to output a set of bounding box sets $B = \{B_1,B_2,\dots B_n\}$ where $B_i$ denotes the set of output bounding boxes for the class $c_i$. These bounding boxes are then counted to obtain the count per cell class.

A possible solution for the aforementioned problem is an object detector $D_\theta$ (see Fig \ref{fig:Model diagram}), which performs both bounding box detection ($B=D_\theta(im))$ and classification. Given ground-truth training data, object detectors are trained with a combination of objectness, classification, and localization losses. Objectness loss ($\mathcal{L}_{obj}$) is the confidence score indicating whether the box contains an object or not.
Classification loss ($\mathcal{L}_{cls}$) is computed as cross-entropy between the predicted class and ground truth class. 
Localization loss ($\mathcal{L}_{loc}$) is the error in predicted bounding box coordinates as compared to ground truth bounding box coordinates. The total detection loss is given by Eq \ref{eq:object detection}. 
\begin{equation}
    \label{eq:object detection}
    \mathcal{L}_{det} = \mathcal{L}_{obj} + \mathcal{L}_{cls} + \mathcal{L}_{loc}
\end{equation}

While object detectors are feasible for MC2DC, in practice, they get bogged with issues such as the presence of limited annotated data, tiny objects, and multiple cell types. To address this, we propose a novel model architecture (\sys), which adds additional components and loss terms, and helps in training a more robust $D_\theta$. 

In particular, \sys{} utilizes \emph{explicit} cell discriminative features (e.g., intensity and size for EN vs IEL) by comparing the distributions of these features over the ground truth and the predicted bounding boxes of each cell type. Additionally, \sys{} computes \emph{implicit} feature embeddings for each bounding box, by training an encoder $E_\phi$. It takes as input an image patch corresponding to a cropped out bounding box $b$ to generate an embedding vector $E_\phi(b)$. Using a supervised contrastive (SupCon) loss for learning these embeddings ensures that they are well separated for different class types. It is to be noted that the explicit features are hand-crafted while the implicit features are data-driven -- trained without any prior knowledge.


\sys{} uses both types of features to fit a Gaussian mixture model (GMM) defined by $G_{\psi}$ for both ground truth and predicted bounding boxes. As shown in the Fig \ref{fig:Model diagram} (Posterior Regularization), it samples from the learned GMM model $G_{\psi}$ and imposes similarity between the predicted and ground truth distributions via the Kullback-Leibler (KL) divergence loss between them -- we call this the \sys{} loss.
\sys{} loss is added to $\mathcal{L}_{det}$ and backpropagated to update the parameters $\theta$. $E_\phi$ is pretrained using SupCon over gold bounding boxes, along with data augmentation and balanced subsampling of classes. 




\subsection{Deep Guided Posterior Regularization }

\sys{} encourages discrimination between cell classes via differences in (explicit or implicit) features. Given a feature $f_j$, our method first computes the average feature value  ($\mathcal{A}_{f_j}(c)$) for each class $c$, over the bounding boxes ($B_c$) of that class. For a pair of classes, \sys{} then computes the difference in these average values ($\mathcal{D}_{f_j}$). All feature differences are concatenated to form vectors ($\mathcal{D}_{F}$), over which GMMs are fit. Finally, we use KL-divergence between the GMMs fits of the true and predicted bounding boxes.

Formally, let $F = \{f_1,f_2, \dots f_m\}$ be a set of $m$ features (implicit and explicit), where $f_{j}(b)$ denotes the value of $j^\mathrm{th}$ feature computed from a bounding box $b$. The average feature value of $f_j$ for the class $c$ is computed as:
\begin{equation}
    \label{eq:Average_feature_value}
    \mathcal{A}_{f_j}(c) = \frac{1}{\lvert B_c \rvert}\sum_{b\in B_c}f_{j}(b)
\end{equation}
Here, $B_c$ is restricted to the bounding boxes for class $c$ in a given image. For this image, the discriminative feature value $\mathcal{D}_{f_j}(c_i,c_k)$ for two classes $c_i$ and $c_k$ is defined as the difference of their average $f_j$ values:
\begin{equation}
    \label{eq:Discriminative_feature_value}
    \mathcal{D}_{f_j}(c_i,c_k) = \mathcal{A}_{f_j}(c_i) - \mathcal{A}_{f_j}(c_k)
\end{equation}
\sys{} concatenates the discriminative feature values corresponding to different features to form a discriminative feature vector denoted by $D_{F}(c_{i}, c_{k})$:
\begin{equation}
   \label{eq:Average_total_feature_value}
    \mathcal{D}_{F}(c_i,c_k) =[\mathcal{D}_{f_1}(c_i,c_k); \ \mathcal{D}_{f_2}(c_i,c_k);\dots \ ;\mathcal{D}_{f_m}(c_i,c_k)]
\end{equation}
Note, that each image in the dataset will have a discriminative feature vector corresponding to it. 
Once \sys{} has the set of discriminative feature vectors ($D_{F}$) for the entire minibatch, it learns the underlying feature distribution using a density estimator. We use Gaussian Mixture Models (GMM) for estimating the densities as they are known to be  universal density approximators. 

Two separate GMMs are learned for each of the ground truth and predicted bounding boxes/classes.
That is, for every pair of classes $c_i,c_k$, we have a GMM $P_{gt}$, which models the discriminative feature vector $\mathcal{D}_F$ of the ground truth bounding boxes and another GMM $Q_{pd}$, similarly, for predicted bounding boxes. The goal is to `align' these two feature vector distributions. \sys{} does this via a minimization of the KL divergence measure, given by Eq \ref{KL_divergence}.
\begin{equation}
    \label{KL_divergence}
    D_{KL}(P_{gt}||Q_{pd}) = \int_{\mathcal{X}}P_{gt}(x) \ln \frac{P_{gt}(x)}{Q_{pd}(x)} dx
\end{equation}
Here, $\mathcal{X}$ represent the space of all features and $x\in\mathcal{X}$ are the individual feature vectors. 
\sys{} uses Monte Carlo estimates \cite{hershey2007approximating} to approximate the integral in Eq. \ref{KL_divergence}  to get an estimate of the KL divergence using Eq \ref{eq: MC_KL}. To do this, it treats the feature vectors obtained from the ground truth and predicted bounding boxes as samples of distributions $P_{gt}$ and $Q_{pd}$, respectively. Let $x_{g}$ and $x_{p}$ be the ground and predicted feature vectors for image $im$, then KL-divergence is approximated as:
\begin{equation}
    \label{eq: MC_KL}
    D_{MC} = \sum_{im}\log(P_{gt}(x_{g})) - \log(Q_{pd}(x_{p}))
\end{equation}
Here, the sum is over all images $im$ in the dataset. 
By the law of large numbers, $D_{MC}$ converges to $D_{KL}$ as number of samples $\to\infty$ \cite{hershey2007approximating}. Hence, $D_{KL} \approx D_{MC}$.


\sys{} computes a loss term for each pair of classes $c_i, c_k$ and then normalises them by the number of pairs, which is $\binom{n}{2}$.  The final \sys{} loss $\mathcal{L}_F$ is calculated as:
\begin{equation}
\label{eq:size intensity loss}
    \mathcal{L}_{F} = \frac{1}{\binom{n}{2}}  \sum_{i=1}^{n-1}\sum_{k = i+1}^{n} D_{KL}(P_{gt(i,k)}||Q_{pd(i,k)})
\end{equation}
We now describe our feature extraction process.

\subsection{Explicit Features in \sys}

Explicit features are hand-crafted to help discriminate between cell classes. 
Here, we use two such features, size ($f_S$) and intensity ($f_I$), nevertheless, \sys{} can also work with any other explicit features.
Each explicit feature is modeled as a scalar (e.g., intensity has 1 scalar value). 

Let $(w_L,h_L)$ and $(w_R,h_R)$  be the top left  and bottom right pixel coordinates of a bounding box $b$, respectively. Then, we define the size ($f_{S}(b)$) of the bounding box as: 
\begin{equation}
    \label{eq:size_feature}
    f_{S}(b) = (w_{R} - w_{L})*(h_{R} - h_{L})
\end{equation}
Similarly, if $I(w,h)$ is the pixel intensity at $(w,h)$, we define the intensity feature ($f_I$) as:
\begin{equation}
    \label{eq:Average_intensity_feature}
    f_{I}(b) = \frac{\sum_{h=h_L}^{h_R}\sum_{w=w_L}^{w_R}I(w,h)}{f_S(b)}
\end{equation}




With the size and intensity features computed as above, $\mathcal{D}_{f_{I}}(c_i,c_k)$ and $\mathcal{D}_{f_{S}}(c_i,c_k)$ from Eq \ref{eq:Average_feature_value} and \ref{eq:Discriminative_feature_value} are used to obtain the explicit feature discriminative feature vectors:
\begin{equation}
    \label{eq:combined_features}
    \mathcal{D}_{F_{I,S}} = [\mathcal{D}_{f_{I}}(c_{i}, c_{k}); \  \mathcal{D}_{f_{S}}(c_{i}, c_{k})]
\end{equation}

Subsequently, we fit GMMs for $\mathcal{D}_{F_{I,S}}$ corresponding to the predicted and ground truth bounding boxes and compute the KL divergence between them, denoted by $\mathcal{L}_{\mathrm{exp}}(c_i,c_k)$. The total explicit posterior regularization loss is given by adding these KL divergences for all pair of classes:
\begin{equation}
\label{eq:size intensity loss_Det}
    \mathcal{L}_{\mathrm{exp}} = \frac{1}{\binom{n}{2}}  \sum_{i=1}^{n-1}\sum_{k = i+1}^{n} \mathcal{L}_{\mathrm{exp}}(c_i,c_k)
\end{equation}

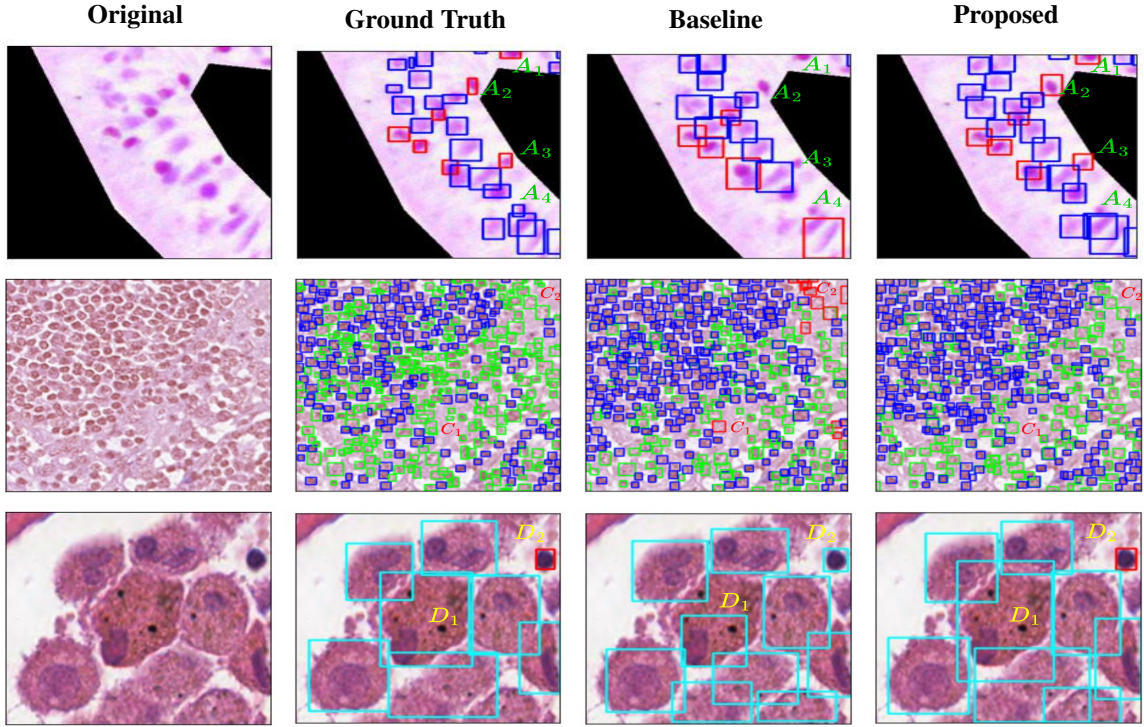
\begin{figure*}
\centering
\begin{tikzpicture}[scale=1.0,transform shape, picture format/.style={inner sep=0.5pt}]

  \node[picture format]                   (A1)   at (0,0)            {\includegraphics[width=1.5in, height = 1.2in]{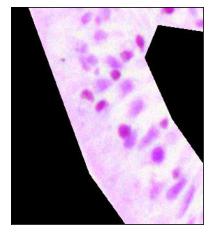}};
  \node[picture format,anchor=north]      (C1) at (A1.south) {\includegraphics[width=1.5in, height = 1.2in]{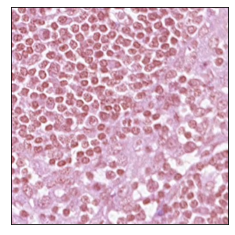}};
  \node[picture format,anchor=north]      (D1) at (C1.south) {\includegraphics[width=1.5in, height = 1.2in]{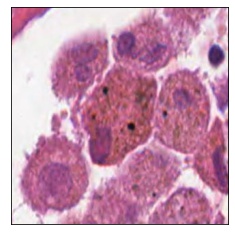}};
  
  \node[picture format,anchor=north west]                   (A2)   at (A1.north east)       {\includegraphics[width=1.5in, height = 1.2in]{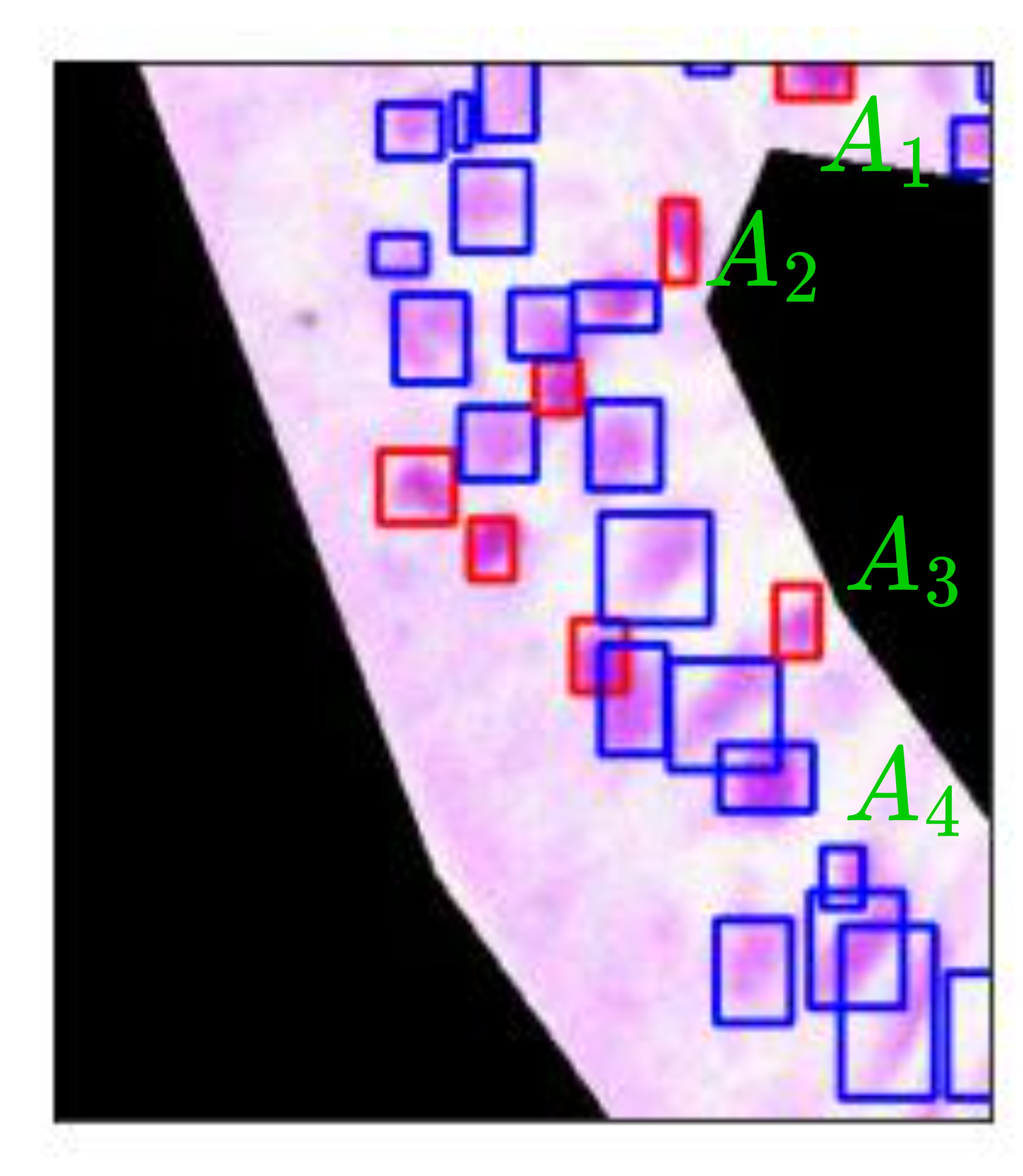}};
  \node[picture format,anchor=north]      (C2) at (A2.south) {\includegraphics[width=1.5in, height = 1.2in]{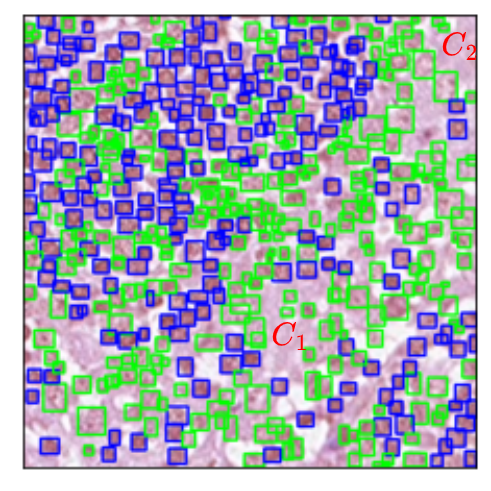}};
  \node[picture format,anchor=north]      (D2) at (C2.south) {\includegraphics[width=1.5in,  height = 1.2in]{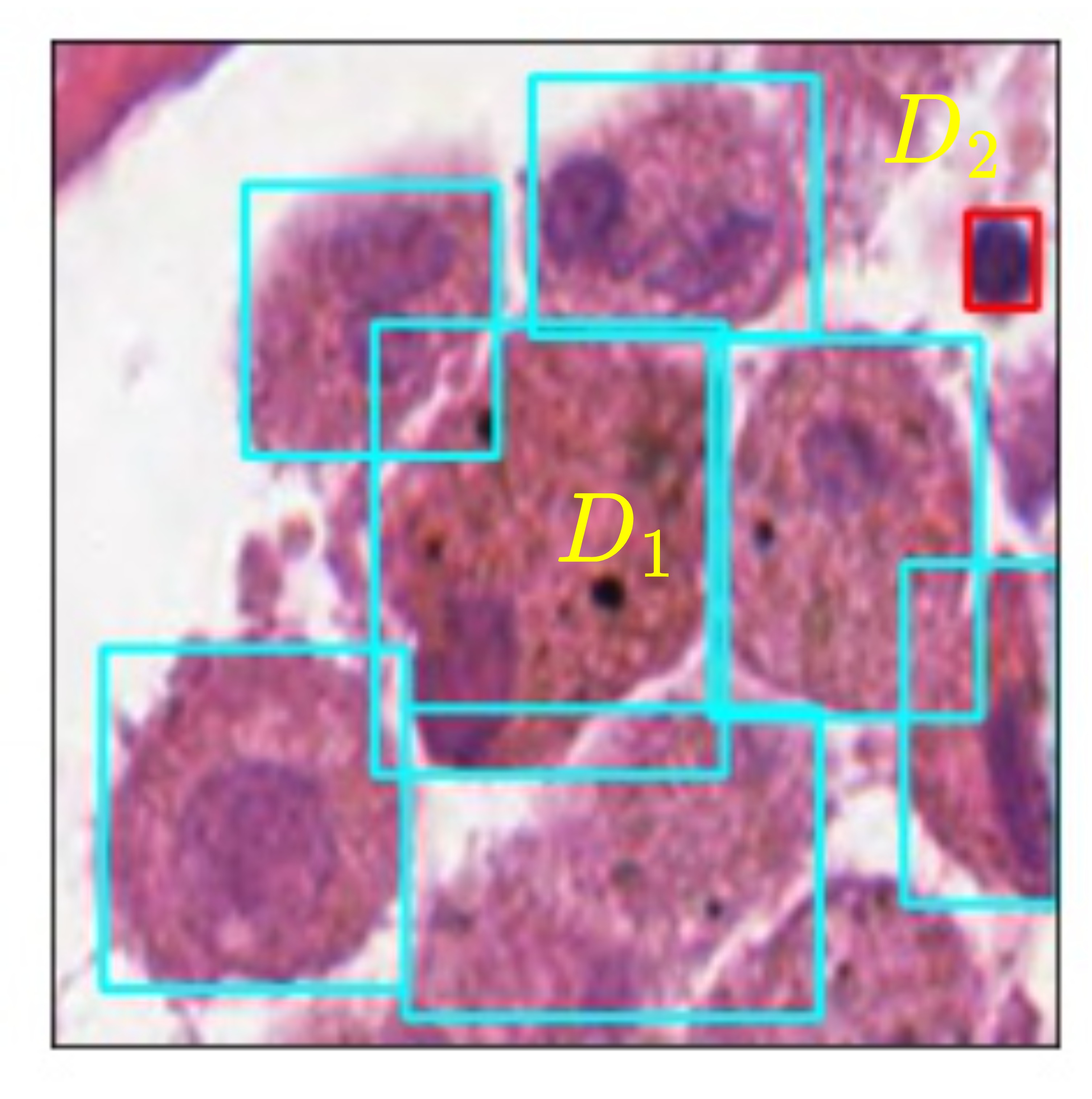}};

  \node[picture format,anchor=north west] (A3) at (A2.north east) {\includegraphics[width=1.5in, height = 1.2in]{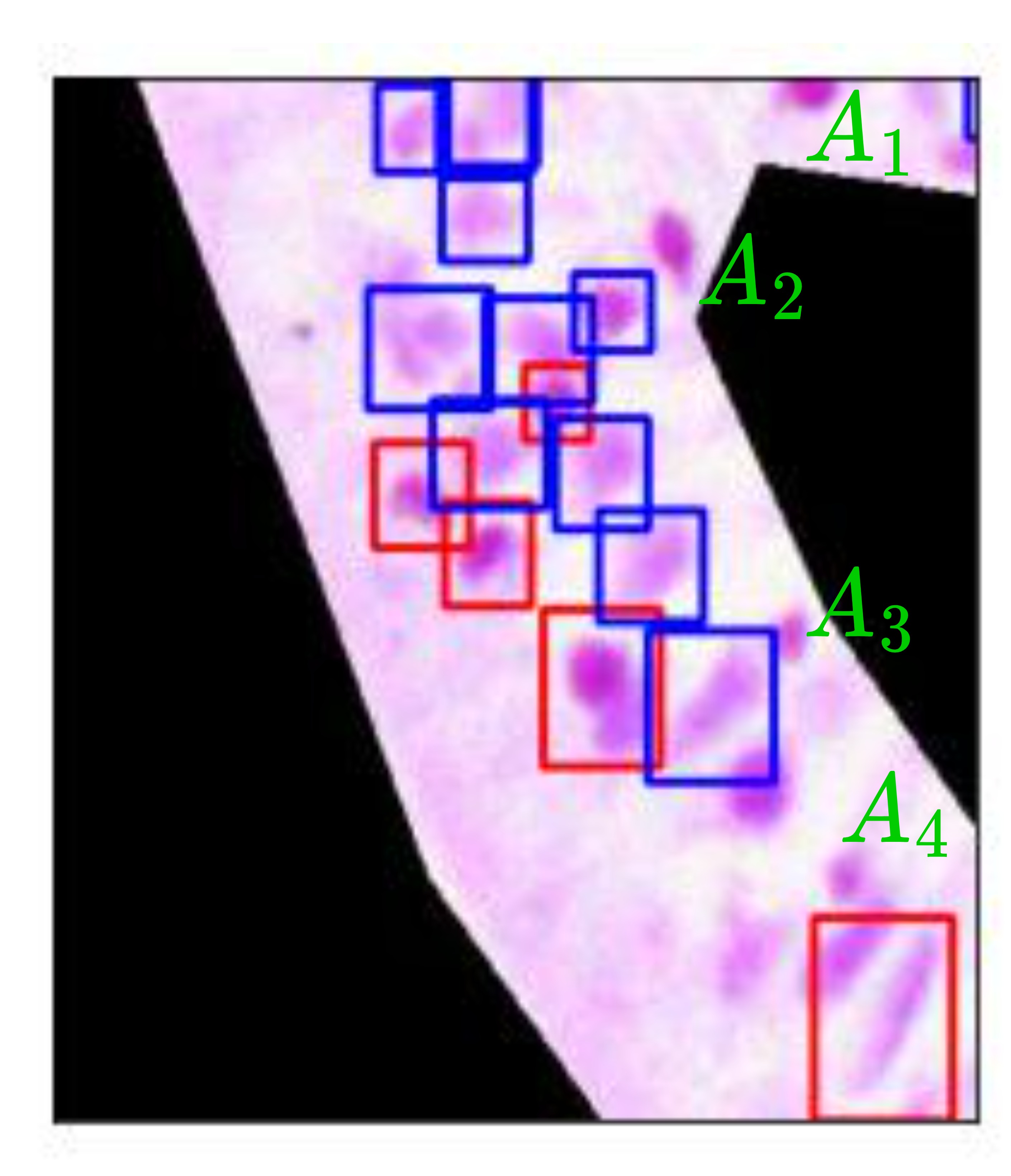}};
   \node[picture format,anchor=north]      (C3) at (A3.south)      {\includegraphics[width=1.5in, height=1.2in]{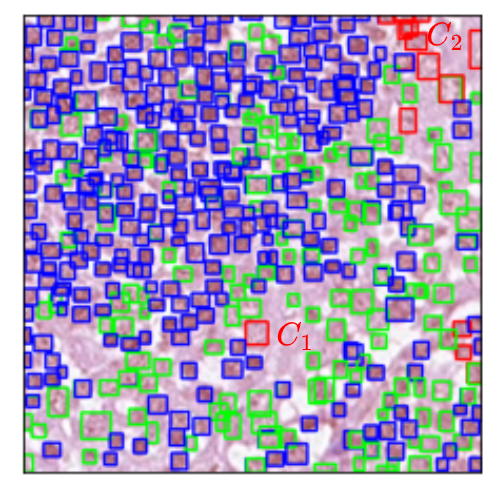}};
   \node[picture format,anchor=north]      (D3) at (C3.south)      {\includegraphics[width=1.5in, height = 1.2in]{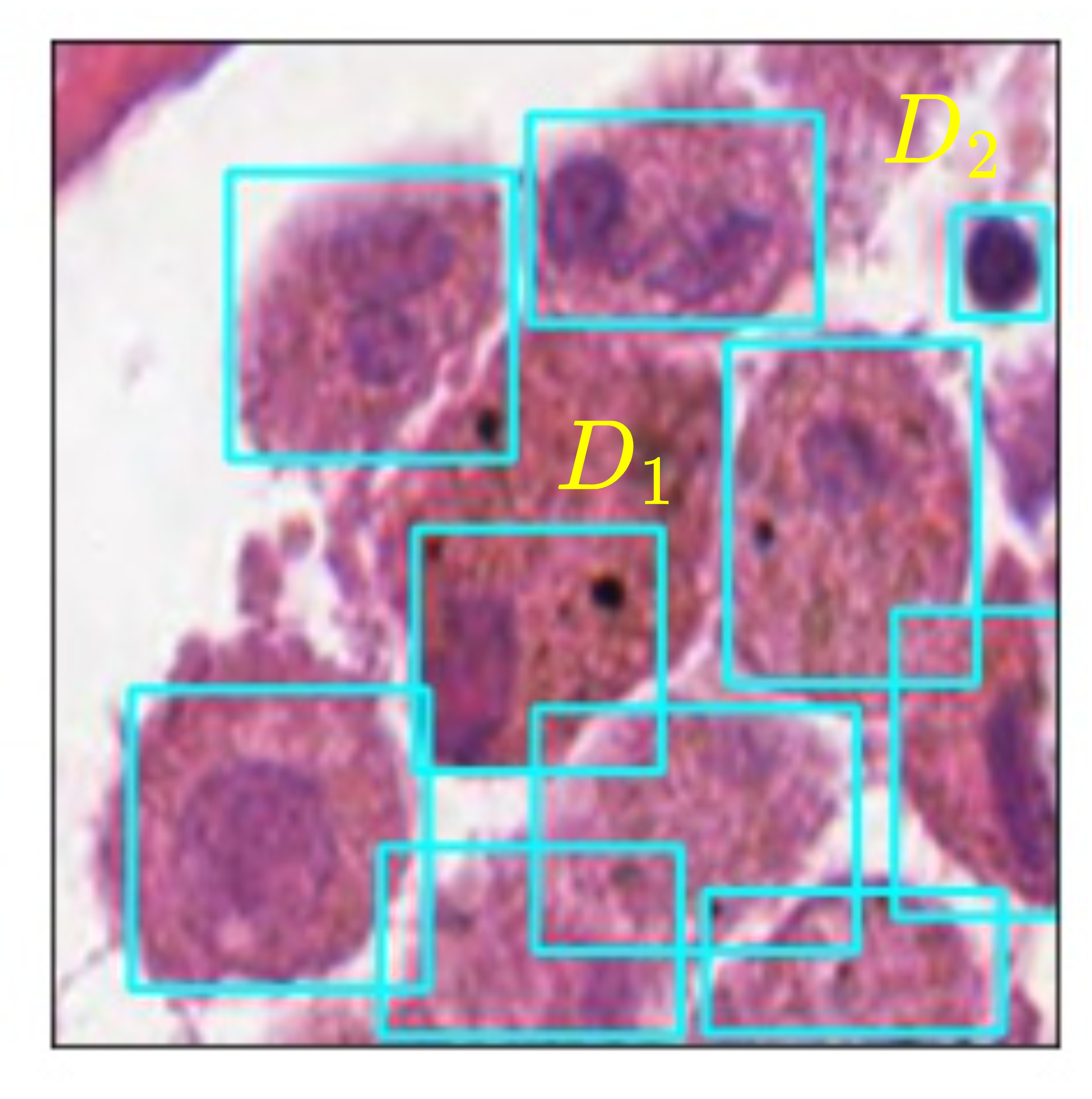}};

  \node[picture format,anchor=north west] (A4) at (A3.north east) {\includegraphics[width=1.5in, height = 1.2in]{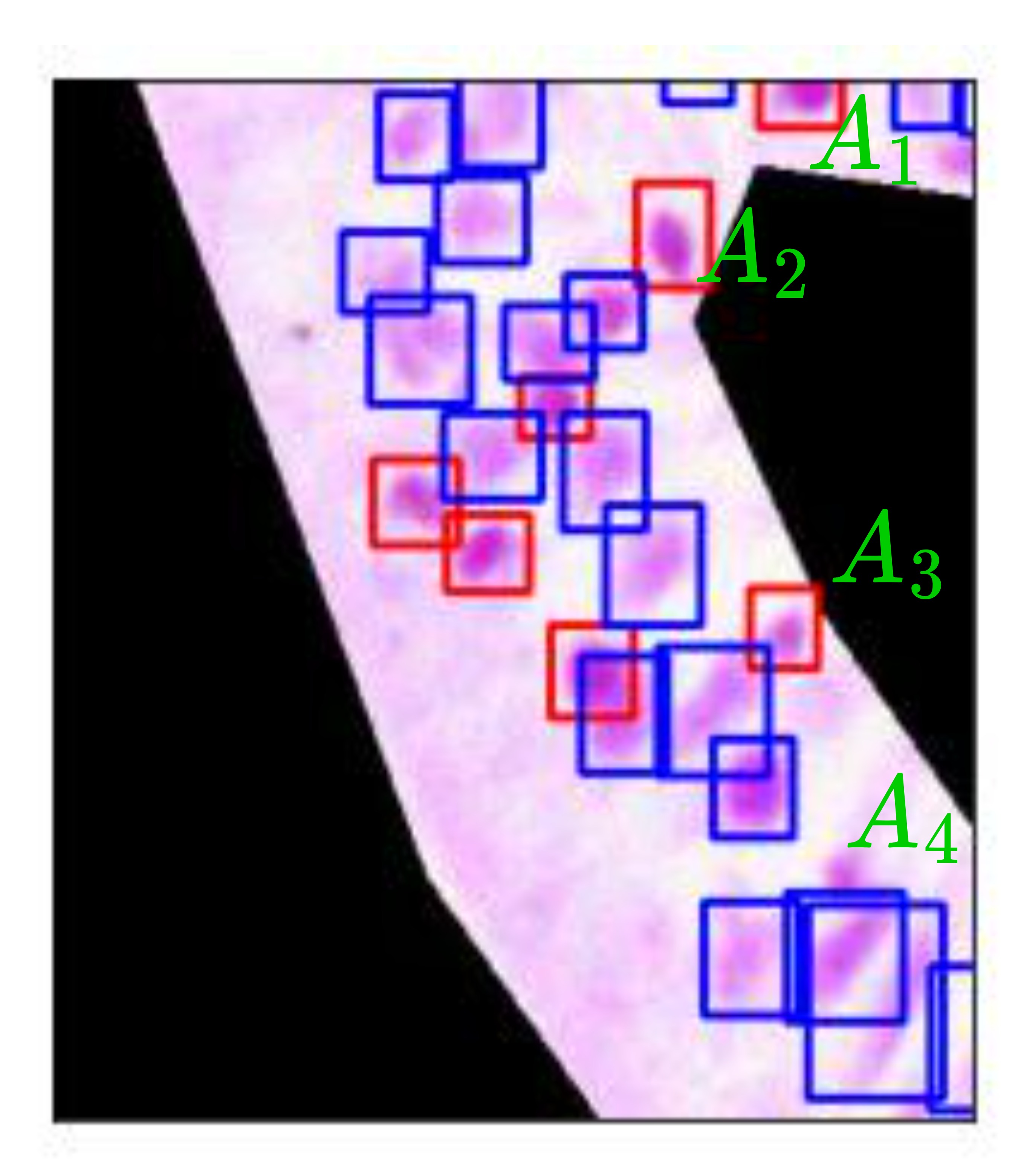}};
  \node[picture format,anchor=north]      (C4) at (A4.south)      {\includegraphics[width=1.5in, height=1.2in]{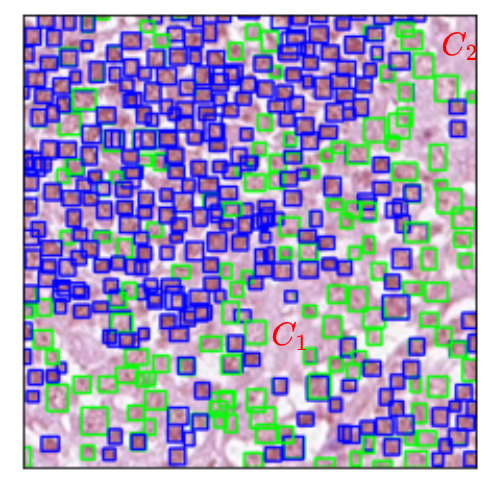}};
  \node[picture format,anchor=north]      (D4) at (C4.south)      {\includegraphics[width=1.5in, height = 1.2in]{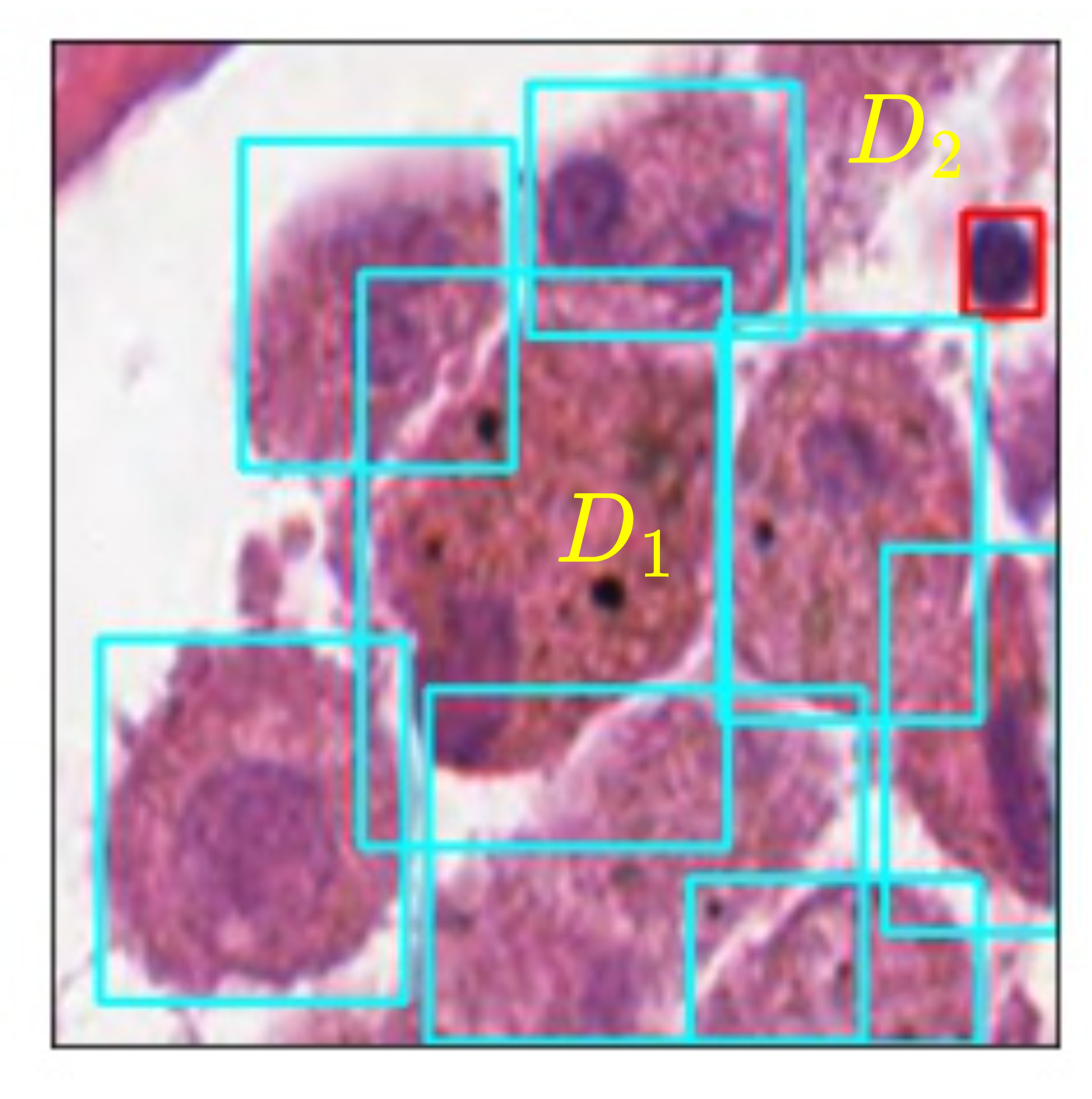}};

  \node[anchor=south] (C1) at (A1.north) {\bfseries Original};
  \node[anchor=south] (C2) at (A2.north) {\bfseries Ground Truth};
  \node[anchor=south] (C3) at (A3.north) {\bfseries Baseline};
  \node[anchor=south] (C4) at (A4.north) {\bfseries Proposed};

\end{tikzpicture}
\caption{Qualitative performance of \sys{}. The third and fourth columns show Yolov5, with and without \sys{}. The first row is an image from MuCeD, with red and blue bounding boxes corresponding to IELs and ENs. The bounding boxes $A_1,A_2,A_3$ show improvement in detecting missing cells and $A_4$ shows improvements in misclassification. Row two is from the CoNSeP dataset with Inflammatory (red), Epithelial (blue), and Spindle (green) cells. Bounding boxes $C_1$ and $C_2$ show improvements in misclassification. Finally, the fourth row from the MoNuSAC dataset shows Epithelial (red), lymphocyte (blue), Neutrophil (green), and Macrophage (cyan) cells. $D_1$ shows improvement in bounding box prediction, while $D_2$ shows improvement in misclassification.}
\label{fig: bbox_prediction}
\end{figure*}

\subsection{Implicit Features in \sys{}}
The information provided by the experts (pathologists) about the discriminative features of the cell types may not be complete or may be hard to compute as an explicit feature. For instance, a shape-related feature (circular vs elongated) is hard to model, when exact segmentation masks are unavailable.\footnote{We tried edge detection over bounding boxes for computing shape features, but noise in highly zoomed medical images resulted in very poorly detected edges.}
To deal with this, \sys{} adopts implicit feature learning and trains a ResNet18 \cite{he2016deep} encoder $E_\phi$, which converts an input image \emph{patch} $v$ to an implicit feature vector $z_v$. Here, each image patch corresponds to a predicted or ground truth bounding box (see Fig. \ref{fig:Shape_size}). Since it may be difficult to learn a GMM on the ResNet18's 512-dimensional feature embedding $z_v$, \sys{} reduces it to a smaller size (10-22), using Principal Component Analysis (PCA), preserving 90\% of explainable variance. The resulting features are denoted by $F_{imp}$.

Similar to the explicit features, \sys{} computes implicit feature discriminative vectors $\mathcal{D}_{F_{imp}}$ using Eq. \ref{eq:Average_feature_value} and \ref{eq:Discriminative_feature_value}.  Subsequently, the KL divergence between the ground truth and predicted GMM fits of the implicit features, $\mathcal{L}_{imp}(c_{i}, c_{k})$ is calculated for all class pairs, which are averaged to form the total implicit feature loss: 
\begin{equation}
\label{eq:shapeloss}
    \mathcal{L}_{imp} =  \frac{1}{\binom{n}{2}} \sum_{i=1}^{n-1}\sum_{k = i+1}^{n} \mathcal{L}_{imp}(c_{i}, c_{k})
\end{equation}




\vspace{1ex}
\noindent
\textbf{Pre-training of Feature Encoder: } 
The encoder is pre-trained using a supervised contrastive (SupCon) loss \cite{khosla2020supervised}. This operates on a pair of ground truth patches and penalizes the encoder if vectors for the patches of the same class are farther away in terms of a distance metric (such as Euclidean distance) compared to vectors for patches of different classes. SupCon is made hardness-aware by a temperature $\tau$ controling the strength of penalties on hard negative patches\cite{wang2021understanding}. It is defined as follows:
\begin{equation}
    \label{eq:supervised contrastive loss}
    L_{SupCon} = \sum_{v\in V} -\log {\frac{1}{\lvert P(v) \lvert}\sum_{p \in P(v)}\frac{\exp(z_{v} z_{p}/\tau)}{\sum_{a}{\exp(z_{v} z_{a}/\tau)}}}
\end{equation}

Here, $L_{SupCon}$ is computed as a sum over set of all gold image patches $V$. For every patch $v$ with a gold label $c$, a set of positive and negative pairs are defined as follows: $p$ is a positive patch of $v$ when it comes from a set $P(v)$, denoting other patches from class $c$. All the other patches $a \in V$, apart from $p$, form the negative samples. 

With these, the objective of $L_{SupCon}$ is to induce a representation space such that similar (positive) sample pairs are close to each other while dissimilar (negative) pairs are far apart. In our case, \sys{} applies $L_{SupCon}$ on the supervised data, and thus the features learned help in discriminating between different cell-classes in the dataset. 

Owing to the imbalance of cell classes in most of the datasets, we use balanced sampling of patches per class, when creating batches for training. 
Furthermore, since the predicted bounding boxes might not exactly overlap with the ground truth bounding boxes, for encoder robustness, we perform augmentation of the gold patches by randomly shifting and resizing the bounding boxes. Further, inspired by the idea of exposure bias \cite{ranzato2015sequence} methods, we gradually introduce these augmented bounding boxes while training the encoder in an annealing manner. These approaches improve the performance of our encoder, and also of detector. 



\subsection{Loss Function}

The detector $D_\theta$ is trained using a combination of object detection and posterior regularization losses. The latter is the sum of losses due to explicit and implicit features. We control the effect of regularization using $\lambda_{reg}$. We keep the encoder $E_\phi$ frozen when training the detector.
\begin{equation}
    \label{Final Loss}
    \mathcal{L}_{total} = \mathcal{L}_{det} + \lambda_{reg}\left(\mathcal{L}_{exp}+\mathcal{L}_{imp}\right)
\end{equation}

\section{Dataset Details}

\noindent
\textbf{Multi-class Celiac Disease Dataset: }
We release MuCeD, a dataset that is carefully curated and validated by expert pathologists.
The H$\&$E-stained histopathology images of the human duodenum in MuCeD are captured through an Olympus BX50 microscope at 20$\times$ zoom using a DP26 camera with each image being  1920$\times$2148 in dimension. The dataset has 55 images, with bounding boxes for 2,090 IELs and 6,518 ENs annotated
using the LabelMe software and are further validated by multiple pathologists. These cells are selected from the epithelial area -- a region of interest that has been explicitly segmented by experts. The epithelial area denotes the area of continuous villi and is used for cell detection, whereas rest of the area is masked out. Further, each image is sliced into 9 subimages and each subimage is re-scaled to 640x640, before it is given as input to object detection models. 
We divide 55 images into five folds of 11 images each 
and report 5-fold cross-validation numbers. Within 44 training images in a given fold, 8 are used for validation and 36 for training. 

\vspace{1ex}
\noindent
\textbf{CoNSeP Dataset: }
To show the effectiveness of \sys{}, we further validate our method on two publicly available datasets. Colorectal nuclear segmentation and phenotypes (CoNSeP) \cite{graham2019hover} is a nuclear segmentation and classification dataset of H\&E stained histology images. Each image is of 1000$\times$1000 dimension and taken at 40$\times$ magnification. The dataset deals with single cancer, colorectal adenocarcinoma (CRA), images. It consists of a total of 41 whole slide images (WSI), which have a total of 24,319 annotated cells of 3 classes: inflammatory cells, epithelial cells, and spindle cells. A total of 27 images are used for training and the rest 14 images are used for testing purposes. 
Since CoNSeP is originally a segmentation dataset, to use it for MC2DC, we preprocess it by converting each segmentation mask into a bounding box. 
Further, we split the 1000$\times$1000 images into 4 subimages of dimension 500$\times$500. This results in an MC2DC dataset of 108 train and 56 test images.

\vspace{1ex}
\noindent
\textbf{MoNuSAC Dataset:}
We similarly use the multi-organ nuclei segmentation and classification (MoNuSAC) challenge dataset \cite{verma2020multi} by preprocessing segmentation masks into bounding boxes. 
MoNuSAC is a large dataset of nucleus boundary annotations and class labels. The dataset has over 46,000 nuclei from 37 hospitals, 71 patients, four organs, and four nuclei types. A total of 209 images (of 46 patients) are used for training and 85 images are used for testing. There are four nuclei types: epithelial nuclei, lymphocytes, neutrophils, and macrophages. 
Each cell type is different in structure and shape from the others. This makes the dataset perfect for our use case. The images are of variable size and we resize them to 640$\times$640, for uniformity. Cells marked as ambiguous are filtered out from evaluation. 

\section{Experimental Setting}
Through our experiments, we wish to answer the following research questions. (1) Is \sys{} model agnostic, i.e., can it be used effectively with multiple object detection models? (2) How much does \sys{} improve the cell detection and counting performance? And, (3) what are the incremental contributions of each of the various model components, such as implicit features, explicit features, and balanced training of the encoder?


\begin{table*}[h!]
\caption{Detection and counting results for MuCeD}
\centering
\setlength{\tabcolsep}{5pt}
\begin{tabular}{p{100pt} p{35pt} p{35pt} p{35pt} p{40pt} p{40pt} p{55pt} p{50pt}}
\hline
Model& Precision& Recall& mAP& MAE IEL& MRE IEL& MAE Epith& MRE Epith\\
\hline
Yolov5&0.711&0.723&0.751&8.97&42.62&14.61&13.43\\
Yolov5 (\sys)&\textbf{0.744}&\textbf{0.735}&\textbf{0.787}&\textbf{5.83}&\textbf{24.19}&\textbf{13.15}&\textbf{12.46}\\
Faster-RCNN&0.592&0.436&0.496&11.85&50.05&27.50&24.93\\
Faster-RCNN (\sys)&\textbf{0.646}&\textbf{0.468}&\textbf{0.541}&\textbf{9.61}&\textbf{31.64}&\textbf{26.50}&\textbf{23.60}\\
EfficientDet&0.266&0.640&0.414&20.35&133.91&20.30&20.78\\
EfficientDet (\sys{})&\textbf{0.274}&\textbf{0.641}&\textbf{0.425}&\textbf{17.32}&\textbf{90.04}&\textbf{18.51}&\textbf{18.12}\\
\hline

\end{tabular}
\label{tab:Celiac_Disease}
\end{table*}

\begin{table*}[h!]
\caption{Detection and counting results for CoNSep}
\centering
\setlength{\tabcolsep}{5pt}
\begin{tabular}{p{100pt} p{35pt} p{35pt} p{35pt} p{50pt} p{50pt} p{55pt} p{50pt}}
\hline
Model& Precision& Recall& mAP& MAE Inflm& MAE Epith& MAE Spindle& MAE Avg\\
\hline
Yolov5&0.638&0.574&0.606&28.21&55.50&57.93&47.21\\
Yolov5 (\sys)&\textbf{0.667}&\textbf{0.584}&\textbf{0.625}&\textbf{26.35}&\textbf{55.00}&\textbf{53.85}&\textbf{45.07}\\
Faster-RCNN&0.490&0.208&0.342&64.71&227.93&198.29&163.64\\
Faster-RCNN (\sys)&\textbf{0.571}&\textbf{0.331}&\textbf{0.451}&\textbf{51.93}&\textbf{151.28}&\textbf{163.00}&\textbf{122.07}\\
EfficientDet&0.633&0.178&0.205&86.00&79.86&134.36&100.27\\
EfficientDet (\sys{})&\textbf{0.672}&\textbf{0.194}&\textbf{0.229}&\textbf{79.64}&\textbf{77.78}&\textbf{125.85}&\textbf{94.42}\\
\hline

\end{tabular}
\label{tab:ConSep}
\end{table*}

\begin{table*}[h!]
\caption{Detection and counting results for MoNuSac}
\centering
\setlength{\tabcolsep}{5pt}
\begin{tabular}{p{95pt} p{35pt} p{35pt} p{35pt} p{50pt} p{50pt} p{50pt} p{50pt}}
\hline
Model& Precision& Recall& mAP& MAE-Epithelial& MAE-Lymphocyte& MAE-Neutrophil & MAE-Macrophage\\
\hline
Yolov5&0.611&\textbf{0.497}&0.481&25.15&14.12&1.96&3.95\\
Yolov5 (\sys)&\textbf{0.736}&0.474&\textbf{0.489}&\textbf{12.01}& \textbf{10.69}& \textbf{0.81}&\textbf{2.33}\\
Faster-RCNN&0.570&0.310&0.405&\textbf{19.52}&23.48&1.0&3.38\\
Faster-RCNN (DeGPR)&\textbf{0.643}& \textbf{0.370}& \textbf{0.473}&19.81&\textbf{22.44}&\textbf{0.82}&\textbf{3.02}\\
EfficientDet&0.256&\textbf{0.509}&0.402&17.67&17.25&1.24&6.51\\
EfficientDet (\sys{})&\textbf{0.258}& 0.499& \textbf{0.409}&\textbf{14.84}&\textbf{16.98}&\textbf{0.56}&\textbf{3.97}\\
\hline

\end{tabular}
\label{tab:MoNuSac}
\end{table*}

\vspace{0.5ex}
\noindent
\textbf{Evaluation Metrics:}
We use precision, recall, and mean average precision (mAP) as the metrics for cell detection. 
For cell counting, we use MAE (mean absolute error) and MRE (mean relative error) as evaluation metrics. MAE provides the absolute difference between predicted count and true counts. MRE provides the relative difference with respect to the true counts. We compute MAE and MRE for the original complete image rather than subimages. 

Additionally, we use the Q-histology \cite{das2019quantitative} parameter for the quantitative classification of duodenum biopsy images into the celiac or non-celiac category. Q-Histology ratio is defined as the ratio of the number of IELs per 100 ENs. If the ratio is $\geq$ 25, then the patient suffers from celiac disease. We use this ratio to evaluate our model on the downstream task of classifying patients into celiac and non-celiac.

\vspace{0.5ex}
\noindent
\textbf{Baselines \& Implementation Details:}
For MuCeD, we pretrain Yolov5 on the Kaggle data science bowl 2018 dataset,\footnote{\tt https://www.kaggle.com/c/data-science-bowl-2018} which is a cell nuclei segmentation challenge, after converting the segmentation masks into bounding boxes. Yolov5 is trained for 300 epochs using SGD optimizer with a learning rate (lr) of 0.003, early stopping with patience 100 and batch size 32. 
We use Faster-RCNN with a Resnet50 backbone. We train Faster-RCNN for 200 epochs with the SGD optimizer and lr of 0.005, weight decay 0.0005 and lr scheduler with step size 3. Finally, for EfficientDet, we use the pretrained Efficientdet-d0 as the base model. We train EfficientDet for MuCeD with a lr of 0.001 for 2000 epochs with patience 100 and is trained with momentum optimizer \cite{sutskever2013importance}. For CoNSep and MoNuSac, EfficientDet is trained with a lr of 0.008, and Faster-RCNN and Yolov5 with 0.03 with lr scheduler with step size 3. All  hyperparameters are fine-tuned using grid search on the respective validation sets. We conduct our experiments using NVIDIA-RTX 5000 and Tesla V100 GPUs. 

While training \sys{}, we use $10^{5}$ samples to approximate KL divergence in Eq \ref{KL_divergence} to get Eq \ref{eq: MC_KL}. We perform a grid search to determine the best regularization factor $\lambda_{reg}$ as $\lambda_{reg} = 0.01$ for MuCeD and CoNSep datasets, and $\lambda_{reg} = 0.001$ for MoNuSac. We also do grid search over relative weights of $\mathcal{L}_{exp}$ and $\mathcal{L}_{imp}$ and observe that 1:1 works best.
All cell patches input to $E_\phi$ are cropped out from bounding boxes and resized to 224$\times$224. Pre-training of ResNet18 encoder is done for 
300 epochs with a lr of 0.001 and momentum \cite{sutskever2013importance} as an optimizer. 
For MuCeD dataset we observe that the model performs the best, when the IoU threshold is kept at 0.3. Hence, MuCeD experiments are performed for mAP:0.3. For CoNSeP and MoNuSAC, we use the standard IoU threshold of 0.5.
We use horizontal flip, vertical flip, scaling and shifting as augmentation methods (more details in appendix).

\section{Results}

\begin{table*}[h!]
\caption{Ablation on MuCeD with Yolov5 baseline}
\label{Ablation_table}
\setlength{\tabcolsep}{3pt}
\begin{tabular}{P{25pt} P{35pt} P{35pt} P{35pt} P{35pt} P{35pt} P{40pt} P{45pt} P{40pt} P{48pt} P{48pt}}
\hline
Yolov5& explicit & implicit & Balance&  Precision& Recall& mAP& MAE IEL& MRE IEL& MAE Epith& MRE Epith \\
\hline
\checkmark&&&&0.711&0.723&0.751&8.97&42.62&14.61&13.43\\
\checkmark&\checkmark&&&0.737&0.723&0.779&\textbf{5.79}&\textbf{23.13}&13.43&13.49\\
\checkmark&&\checkmark&&0.724&0.735&0.771&5.83&23.50&12.98&13.89\\
\checkmark&\checkmark&\checkmark&&0.741&0.734&0.780&6.06&23.40&\textbf{12.57}&13.05\\
\checkmark&\checkmark&\checkmark&\checkmark&\textbf{0.744}&\textbf{0.735}&\textbf{0.787}&5.83&24.19&13.15&\textbf{12.46}\\
\hline
\end{tabular}
\label{tab:Ablation}
\end{table*}

\noindent
\textbf{Detection and Counting Metrics: }
Table \ref{tab:Celiac_Disease} shows the quantitative performance of \sys{} on MuCeD dataset. We compare the object detection models with and without \sys{}. We notice that there is a substantial improvement in all metrics and over all baselines, when \sys{} is used. It suggests that the guidance provided through explicit and implicit features helps the detection model to learn discriminating attributes for the cells. We particularly note that relative error for IEL counts (the minority class) has a drastic reductions of 18-43\% points, showing the effectiveness of the approach.
We observe a similar trend (tables \ref{tab:ConSep} and \ref{tab:MoNuSac}) on CoNSeP and MoNuSAC datasets. While \sys{} performance is stronger than the baselines in all settings, we note that counting results in CoNSep are generally weak for all models -- we suspect this is because the density of cells in that dataset is quite high (585 cells/image, compared to 70 and 156 for MoNuSac, MuCeD, resp.), and models end up missing a fraction of cells, leading to high absolute errors. 

Qualitatively, we illustrate model predictions in Fig \ref{fig: bbox_prediction}. The first column depicts the original image, the second is ground truth bounding boxes, the third  shows the image with predictions from the Yolov5 baseline model, while the final column shows predictions from the Yolov5 with \sys{}. Three rows contain an exemplar image each from MuCeD, CoNSeP, and MoNuSAC, respectively. We note that \sys{} reduces both misclassification and misidentification errors. The highlighted bounding boxes $A_{1}, A_{2}, A_{3}$ show improvement in the detecting missing cells, and $A_{4}, C_{1}, C_{2}, D_{2}$ show reductions in misclassification.

We further classify the patient samples in MuCeD, based on the Q-histology ratio. Table \ref{tab:Q_Classification_metric} reports the comparative analysis. \sys{} improves prediction accuracy from 
74.55\% to 87.7\% and celiac F1-score from 0.774 to 0.902. 

\begin{table}[t]
\caption{Classification Metrics based on Q-Ratio}
\centering
\setlength{\tabcolsep}{4pt}
\begin{tabular}{p{75pt} p{38pt} p{55pt} }
\hline
Measure & Baseline & Yolo+\sys{} \\
\hline

Recall & 0.774 & \textbf{0.936}\\
Precision& 0.774 & \textbf{0.871}\\
F1-score& 0.774 & \textbf{0.902}\\
Accuracy& 0.746 & \textbf{0.877}\\
\hline
\end{tabular}
\label{tab:Q_Classification_metric}
\end{table}

\vspace{0.5ex}
\noindent
\textbf{Comparison against Other Counting Models: }
Tables \ref{tab:Counting_vs_Localisation_MuCeD} and \ref{tab:Counting_vs_Localisation_CoNSep} compare the performance of counting via detection (Yolov5+\sys{}) vs density map based methods. For comparison, we use four state-of-the-art counting models: UNet \cite{ronneberger2015u}, FRCN-A \cite{xie2018microscopy}, Count-ception \cite{paul2017count} and SAU-Net \cite{sau-net}. As all these methods expect a single-class input, we train separate models for each class, and aggregate. We observe that our approach outperforms all other methods by vast margins. Also, counting via detection in a multi-class setting is computationally convenient, as we can get the counts of all cell types from a single object detector. We also compare with MCSpatNet \cite{abousamra2021multi} and observe improved performance with \sys{} (see appendix).

\begin{table}[t]
\caption{Counting vs Localization (MuCeD)}
\setlength{\tabcolsep}{4pt}
\begin{tabular}{p{75pt} p{40pt} p{40pt} p{35pt}}
\hline
Model& MAE-IEL& MAE-Epith& MAE-Avg\\
\hline
UNet&11.72&26.85&19.29\\
FCRN-A&15.60&22.81&19.21\\
Countception&16.10&29.78&22.94\\
SAU-Net&11.56&28.07&19.82\\
Yolov5 (\sys{})& \textbf{5.83}& \textbf{13.43}&\textbf{9.63}\\
\hline

\end{tabular}
\label{tab:Counting_vs_Localisation_MuCeD}
\end{table}

\begin{table}[t]
\caption{Counting vs Localization (CoNSeP)}
\setlength{\tabcolsep}{3pt}
\begin{tabular}{p{72pt} p{34pt} p{35pt} p{30pt} p{28pt} }
\hline
Model& MAE-Inflamm& MAE-Epithelial& MAE-Spindle & MAE-Avg\\
\hline
UNet&64.03&101.11&159.47&108.20\\
FCRN-A&53.18&94.34&95.84&81.12\\
Countception&77.13&129.61&151.13&119.29\\
SAU-Net&50.72&77.38&99.14&75.75\\
Yolov5 (\sys{})& \textbf{26.35}& \textbf{55.00}& \textbf{53.85}& \textbf{45.07}\\
\hline
\end{tabular}
\label{tab:Counting_vs_Localisation_CoNSep}
\end{table}

\vspace{0.5ex}
\noindent
\textbf{Ablation Studies: }
We perform ablation analysis to understand the relative contributions of different components in the final performance. We run this study for MuCeD using Yolov5. 
All results are reported in Table \ref{tab:Ablation}. Comparing rows 1 and 2, we notice that explicit features improve precision from 0.711 to 0.737 and mAP from 0.751 to 0.774. We observe a similar performance gain in counting metrics. Similarly, introduction of implicit features (rows 1 vs. 3) improves mAP from 0.751 $\rightarrow$ 0.771 and MAE reduces from 8.79 $\rightarrow$ 5.83 along with improvements in other metrics. We also find that the implicit and explicit features capture complementary information (details in appendix Sec.5).

To mitigate class imbalance while contrastive pre-training of the encoder, we perform class-balanced sampling while creating batches, and additional data augmentations for robustness. Comparing rows 4 and 5, we note small improvements in most metrics, due to these. 

\vspace{0.5ex}
\noindent
\textbf{Error Analysis:} We also perform error analysis for our best model on MuCeD. The common failure modes include more errors in misclassifying IELs (minority class) as ENs than reverse. This is especially true if an IEL (circular) overlaps an EN (elongated), since the overall shape appears elongated. The darker stained images generally produce more errors, presumably because the intensity differences between cell types are diminished. Finally, EN cells are missed when they are very lightly stained.

\section{Conclusions}
We study multi-class cell detection and counting problems (MC2DC) over medical histopathological images in a limited data setting. Our solution, Deep Guided Posterior Regularization (\sys{}), imposes additional regularization terms, incentivizing the model to output, for each cell class, a posterior distribution of features over predicted bounding boxes similar to that of ground truth. \sys{} uses two types of features: explicit -- generally provided by a domain expert, and implicit -- trained automatically using a supervised contrastive loss over labeled data. 

We also contribute a novel dataset of 55 duodenum biopsies (useful for predicting celiac disease) for our task, along with experimenting on two publicly available datasets. 
We find that \sys{} is effective in improving performance of several object detection backbones, obtaining substantial improvements in both detection and counting metrics. As a consequence, the F-score of the model in predicting celiac disease increases from 77\% to 90\%. 
We release our code and data for further research.

\vspace{0.5ex}
\noindent
\textbf{Acknowledgements: }
This work is primarily supported by IITD-AIIMS Multi-Institutional
Faculty Interdisciplinary Research Project and an ICMR grant on automated grading of duodenal mucosal biopsies for celiac disease. It is also supported by Jai Gupta chair fellowship by IIT Delhi, Yardi School of AI for travel and honorarium grants. We thank the IIT Delhi HPC facility for computational resources.

{\small
\bibliographystyle{ieee_fullname}
\bibliography{egbib}
}

\end{document}